\documentclass[review]{elsarticle}

\usepackage{lineno,hyperref}
\usepackage[margin=2.5cm]{geometry}

\usepackage{graphicx}
\usepackage{amsmath}
\usepackage{amssymb}
\usepackage{algorithm}
\usepackage{algorithmic}
\usepackage{subfigure}
\usepackage{mathtools}
\usepackage{threeparttable}
\usepackage{comment}
\usepackage{multirow}
\usepackage{wrapfig}
\usepackage{lipsum,booktabs}
\usepackage{subfigure} 
\usepackage{pifont}
\newcommand{\cmark}{\ding{51}}%
\newcommand{\xmark}{\ding{55}}%

 \usepackage{amsmath}

\newtheorem{proposition}{Proposition}

\journal{Journal of Neural Networks\ Templates}

\begin{document}

\begin{frontmatter}

\title{Sparsity-Control Ternary Weight Networks}




\author[1]{Xiang Deng\corref{mycorrespondingauthor}}
\cortext[mycorrespondingauthor]{Corresponding author}
\ead{xdeng7@binghamton.edu}

\author[]{Zhongfei Zhang}

\address[1]{State University of New York at Binghamton, Binghamton, NY, US}

\begin{abstract}
Deep neural networks (DNNs) have been widely and successfully applied to various applications, but they require large amounts of memory and computational power.
This severely restricts their deployment on resource-limited devices.
To address this issue, many efforts have been made on training low-bit weight DNNs.
In this paper, we focus on training ternary weight \{-1, 0, +1\} networks which can avoid multiplications and dramatically reduce the memory and computation requirements.
A ternary weight network can be consider as a sparser version of the binary weight counterpart by replacing some -1s or 1s in the binary weights with 0s, thus leading to more efficient inference but more memory cost.
However, the existing approaches to training ternary weight networks cannot control the sparsity (i.e., percentage of 0s) of the ternary weights, which undermines the advantage of ternary weights.
In this paper, we propose to our best knowledge the first sparsity-control approach (SCA) to training ternary weight networks, which is simply achieved by a weight discretization regularizer (WDR).
SCA is different from all the existing regularizer-based approaches in that it can control the sparsity of the ternary weights through a controller $\alpha$ and does not rely on gradient estimators.
We theoretically and empirically show that the sparsity of the trained ternary weights is positively related to $\alpha$.
SCA is extremely simple, easy-to-implement, and is shown to consistently outperform the state-of-the-art approaches significantly over several benchmark datasets and even matches the performances of the full-precision weight counterparts.
\end{abstract}


\end{frontmatter}


\section{Introduction}

Deep Neural Networks (DNNs) have obtained state-of-the-art performances on various tasks across a wide range of domains including computer vision \cite{girshick2015fast, redmon2016you, kroner2020contextual,clay2021learning} and natural language processing \cite{serban2016generating, andreas2016learning, joulin2016bag}.
Currently, there is an increasing interest in implementing DNNs on portable devices, driven by the strong and demanding needs in a wide spectrum of applications noticeably including smart sensor networks, telehealthcare, and security.
However, as DNNs need large amounts of memory and computing power, it is still a challenging problem on how to effectively implement DNNs on the devices with constrained memory or limited computing power.\par

One solution to this problem is to reduce the bit-width of DNN weights, such as training DNNs with binary weights or ternary weights.
In this paper, we focus on training ternary weights without considering discrete activations.
Ternary weight networks are different from binary weight networks in that they set some values in the binary weights to zeros, which makes them sparser, thus being more efficient but also requiring more memory.
Ternary weight networks can be divided into three categories, i.e., the ternary weight with values constrained to \{-1, 0, +1\}, the ternary weight with \{-$\alpha$, 0, +$\alpha$\}, and the ternary weight with \{-$\alpha$, 0, +$\beta$\} where $\alpha$ and $\beta$ are real numbers and $\alpha \neq \beta$.
Among these three kinds of ternary weights, the ternary weight with \{-1, 0, +1\} is the most efficient\footnote{\{-$\alpha$, 0, +$\alpha$\} is more efficient than \{-$\alpha$, 0, +$\beta$\} as it can be considered  as \{-1, 0, +1\} $\times \alpha$.} as it avoids multiplications completely, and it is even more efficient than the binary counterpart as ternary weights \{-1, 0, +1\} can be considered as setting some values in the binary weights \{-1, +1\} to 0s.
0s can be considered as network pruning \cite{frankle2018lottery} that makes networks sparse and thus reduces the computational cost.
In light of this, in this paper, we attempt to train DNNs with ternary weights \{-1, 0 +1\} that are able to reduce the computational complexity substantially, speed up the inference significantly, and lead to about 16$\times$ or 32$\times$ memory requirement reduction compared with the float (32-bit) or double (64-bit) precision counterparts. \par
\par


\begin{table}[t]
\caption{Comparison between WDR and the existing regularizers}
\centering
\label{refdif}
\begin{tabular}{lcc}
\toprule
Methods                                                    & Controlling sparsity &Free from gradient estimators \\ \midrule
The regularizer in \cite{tang2017train}   &  \xmark                 &              \xmark                    \\ \midrule
The regularizers in  \cite{darabi2018bnn+} &     \xmark                   &        \xmark                             \\ \midrule
SCA                      &      \cmark                 &     \cmark
\\ \bottomrule
\end{tabular}
\end{table}

The challenge for training ternary or binary weight networks is that the gradients with respect to the discrete weights do no exist.
Substantial efforts have been made on addressing this issue \cite{zhang2018lq, leng2018extremely, zhou2017incremental, xu2019iterative}.
These approaches train discrete weight networks by using stochastic weights or using the straight-through estimator to estimate the gradients with respect to discrete weights.
However, we notice that most existing efforts on training ternary weight networks focus on training ternary weights with \{-$\alpha$, 0, +$\alpha$\} \cite{li2016ternary, leng2018extremely, zhou2018explicit} or \{-$\alpha$, 0, +$\beta$\} \cite{zhu2016trained,zhou2017incremental}, which are unable to avoid multiplications completely.
Most of these approaches still have a large performance gap to full-precision counterparts, while some approaches \cite{shayer2017learning,zhou2017incremental} are able to obtain promising results, but they need a complex training process, which causes the ternary weight networks not widely used.
Moreover, none of these approaches is able to control the sparsity of the ternary weights, which cannot fully use the advantage of ternary weights.
It is thus necessary and appealing for the development of an accurate, sparsity-controlling and easy-to-implement approach for converting full-precision networks to the ternary versions with weights \{-1, 0, +1\}.
\par

In the paper, we propose a simple yet effective approach (i.e., SCA) to training ternary weight \{-1, 0, +1\} networks and it also has the ability to control the sparsity of the weights.
SCA is developed base on the fact that the ternary solutions to a DNN are also included in the full-precision weight space.
SCA attempts to search for a ternary solution in the full-precision space.
However, it is difficult to find an appropriate ternary solution in such a large weight space.
To address this issue, at training time, SCA limits the full-precision weight to be in the range between -1 and +1 by parameterizing them with $\tanh(\Theta)$ where $\Theta$ are the parameters.
In this way, the weight space is largely reduced.
By taking advantage of the properties of the $\tanh(\Theta)$ function, we design a novel weight discretization regularization (WDR) to force weight ternarization and control the sparsity.
To the end, SCA simply trains a ternary weight network by optimizing a full-precision network loss plus a WDR term. 
At test time, the ternary weights are obtained by simply rounding $\tanh(\Theta)$ to the nearest integers.
Despite its simplicity, it is able to outperform the state-of-the-art approaches and perform on a par with full-precision counterparts.
\par

SCA is a regularizer-based approach.
In the existing literature, three well-known regularizers \cite{tang2017train,darabi2018bnn+} have been proposed for training binary weight networks and they can be easily extended to training ternary weight networks.
We summarize the differences between the regularizer (i.e., WDR) in SCA and the existing regularizers in Table \ref{refdif}.
First, SCA is able to \textbf{control the sparsity} of the ternary weights while the existing regularizers cannot.
Second, SCA is able to use the real gradients to train ternary weight networks while the existing regularizers rely on the straight-through gradient estimator.
We provide the technique differences among these regularizers in more detail in Section \ref{difs}.

\par

The main contributions of our work can be summarized as follows:
\begin{itemize}
\item We have proposed a simple yet effective approach (i.e., SCA) to training ternary weight networks by simply minimizing a task loss plus a regularizer.
The simple nature of SCA makes it easy-to-implement and have a high code-reuse-rate when converting a full-precision network to the ternary version, which is of much practical values, considering the fact that the state-of-the-art models in various domains are still built on full-precision weight DNNs.

\item We have proposed a novel regularizer WDR for training ternary weights.
The shape controller $\alpha$ in WDR enables SCA to control the sparsity of the trained ternary weights.
We theoretically and empirically show that the sparsity of the ternary weights is positively related to $\alpha$.

\item
The existing literature has only studied how to control the sparsity of full-precision weights in a DNN.
To the best of our knowledge, this is first work to explore controlling the sparsity of ternary weights in a DNN.

\item Extensive experiments on several benchmark datasets demonstrate that SCA outperforms the state-of-the-art approaches and matches the performances of full-precision counterparts.

%


\end{itemize}

\section{Related Work}
\label{headings}

Our work is related to the literature on model compression including network quantization and network pruning.
Thus, we first present an overview of the existing techniques on training DNNs with discrete weights and then review the existing literature on network pruning.

\subsection{Network Quantization}

Network quantization has drawn numerous research attentions due to its potential in various applications and low memory and computational power requirements.
The goal is to train DNNs with low-bit weights or activations.
These approaches train discrete neural networks by approximating full-precision weights or activations in each layer with scaling factors and discrete values \cite{rastegari2016xnor,li2016ternary,alemdar2017ternary,mellempudi2017ternary,lin2017towards,zhu2016trained,mcdonnell2018training,zhou2018explicit,zhuang2019structured,zhang2018lq,martinez2020training,stock2019and}, using stochastic weights \cite{soudry2014expectation,shayer2017learning,meng2020training,stamatescucritical,shekhovtsov2020path}, using a gradient estimator \cite{esser2019learned,li2019additive,bulat2020high}, using the straight-through estimator \cite{courbariaux2015binaryconnect,hubara2016binarized}, or using reinforcement learning \cite{wang2019learningCIBCNN}.
Among these approaches, our work is most related to these approaches to training binary or ternary weight neural networks.
\par

\subsubsection{Approaches to Training Binary Neural Networks}
Many approaches have been proposed to training binary neural networks.
Soudry et al. \cite{soudry2014expectation} propose to train binary neural networks through the variational Bayesian method that infers networks with binary weights and neurons.
BinaryConnect \cite{courbariaux2015binaryconnect} uses sign function to binarize the weights during the forward and backward propagation while using full-precision weights in the parameter update stage.
Binarized Neural Networks (BNNs) \cite{hubara2016binarized} and XNOR-net \cite{rastegari2016xnor} make some extensions to this method by binarizing both weights and activations. 
BNNs \cite{hubara2016binarized} utilizes the sign function and additional scaling factors to binarize the real-value weights and the pre-activations at training time.
The straight-through estimator is used to back-propagate through the binarization operation. 
XNOR-net takes BNNs one step further by approximating the real-valued tensor and activation tensor by a binary filter and a scaling factor. 
ABC-nets \cite{lin2017towards} make an extension by using the linear combination of multiple binary tensors to approximate full-precision weights.
Moreover, to alleviate information loss, multiple binary activations are also employed. 
McDonnell \cite{mcdonnell2018training} follows BinaryConnect \cite{courbariaux2015binaryconnect} and BNNs \cite{hubara2016binarized}, and applies a layer-dependent scaling to the sign of the weights. 
Tang et al. \cite{tang2017train} explore how the learning rate, the scale factor, and the regularizer influence the performances of BNNs.
Shayer et al. \cite{shayer2017learning} propose LR-Net to train binary weight neural networks by using the central limit theorem and the local reparameterization trick. 
They update the weight distribution at training time.
Binary weights are sampled from the learned distribution at test time.
PBNet \cite{peters2018probabilistic} extends the idea of LR-Net and uses a probabilistic method for training neural networks with both binary weights and binary activations.
Leng et al. \cite{leng2018extremely} propose to train low-bit neural networks by decoupling the continuous parameters from the discrete constraints and casting the original problem into several subproblems.
CI-BCNN \cite{wang2019learningCIBCNN} trains binary neural networks by using the channel graph structure and channel-wise interactions.
LQ-Nets \cite{zhang2018lq} jointly train the network parameters and its associated quantizers for DNNs.
Bi-real nets \cite{liu2018bireal} use the identity shortcut to connect the real activations to activations of the consecutive block, thus improving representational capability of binary neural networks.
ELQ \cite{zhou2018explicit} trains discrete weight DNNs through explicitly regularizing the weight approximation error and the loss perturbation.
Zhuang et al. \cite{zhuang2019structured} propose to divide the network into groups of which each can be reconstructed by using a set of binary branches.
Meng et al. \cite{meng2020training} propose to use the Bayesian rule to train binary weights for DNNs.
These approaches either have a large performance gap to the full-precision networks or need a complex training process.

\subsubsection{Approaches to Training Ternary Neural Networks}
Substantial research efforts have been made on training neural networks with ternary weights or activations. 
Alemdar et al. \cite{alemdar2017ternary} use a two-stage teacher-student approach for training neural networks with ternary activations.
First, they train the teacher network with stochastically firing ternary neurons and then let the student network learn how to imitate the teacher's behavior using a layer-wise greedy algorithm. 
Mellempudi et al. \cite{mellempudi2017ternary} propose to take advantage of a fine-grained quantization technique that involves multiple scaling factors to obtain a ternary neural network.
TTQ \cite{zhu2016trained} uses two full-precision scaling coefficients for each layer and quantizes the weights to three real values.
Li et al. \cite{li2016ternary} develop an approach to training ternary weight networks by minimizing the Euclidean distance between the full-precision weights and the ternary weights along with a non-negative scaling factor.
LR-Net \cite{shayer2017learning} attempts to train a stochastically ternary network by leveraging the local reparametrization trick and sampling ternary weights at the test time. 
We notice that almost all the existing efforts on training ternary weight networks focus on training ternary weights with \{$-\alpha$, 0, $\alpha$\} \cite{li2016ternary, leng2018extremely, mellempudi2017ternary, zhou2018explicit} or \{$-\alpha$, 0, $\beta$\} \cite{zhu2016trained, zhou2017incremental} that are unable to avoid multiplications completely and thus are less efficient than \{-1, 0, +1\} in SCA.
More importantly, none of these approaches is able to control the sparsity of the ternary weights, which cannot fully use the advantage of ternary weights.
\par

SCA trains ternary weight networks by using parameterization and a novel regularizer.
Thus, it is also related to regularizer based approaches for training discrete weight networks.
Tang et al. \cite{tang2017train} and BNN+ \cite{darabi2018bnn+} propose three regularizers for training binary neural networks, which can be easily extended to train ternary weight networks.
However, these regularizers cannot control the sparsity of the ternary weights and highly rely on the straight-through estimator to approximate the gradients with respect to the discrete weights.
SCA overcomes all these shortcomings.

\par


\subsection{Network Pruning}
On the other hand, our work is also related to the studies on network pruning, including but not limited to \cite{denton2014exploiting,han2015deep,hu2016network,he2017channel,luo2017thinet,wang2016cnnpack,zhuang2018discrimination, renda2020comparing}.
Network pruning aims to compress a DNN into a sparse version by making the full-precision weights sparser.
Luo et al. \cite{luo2017thinet} propose to use the statistics of next layer to prune weights. 
However, weight pruning approaches require substantial iterations to converge and the pruning threshold needs to be set manually.
Low-rank decomposition has also been introduced to network compression.
These approaches \cite{denil2013predicting,kim2015compression,ren2018deep} use matrix decomposition
technique to decompose the weight in DNNs.
The limitation is that these approaches increase the number of the layers in a DNN and thus are easy to result in gradient-vanishing issues during the training process.

SCA is different from all these existing studies in that SCA attempts to control the sparsity of ternary weights rather than full-precision weights, which is a more challenging problem.
To our best knowledge, SCA is first work to control the sparsity of ternary weight networks.

\section{Framework}
\label{frame}

In this section, we introduce SCA for training DNNs with ternary weights whose values are constrained to \{-1, 0, +1\}.
SCA is developed based on the fact that the ternary solution to a DNN is also included in the full-precision weight space.
To illustrate the connection between ternary weights and full-precision weights, we first review the basic process for training full-precision weight networks.
Then we make further derivations of the basic process to show how to train ternary weight networks.

\subsection{Deep Neural Networks with Full-Precision Weights}
In this part, we review how to train a DNN with full-precision weights.
Given the training data $(X, Y)$ where $X$ are the inputs and $Y$ are the targets, the output of a neural network $f$ with weights $W$ is $f(X, W)$.
The loss can be written as:
\begin{equation}
\label{1}
L_c(W) = \mathcal{L}(f(X, W), Y)
\end{equation}
where $\mathcal{L}(.)$ is any loss function, such as mean square error or cross entropy, and $W=[w_1, w_2, ..., w_n]$ with $w_i$ representing the weights in the $i$th layer.
\par

With the loss function, we can use backpropagation to compute the gradients with respect to the weights $W$ and utilize gradient descent to minimize the loss function.
This can be done easily for continuous weight networks while it is difficult for discrete weight networks due to no gradients.
Fortunately, the discrete solutions to a DNN are also included in the full-precision weight space.
We show below that with modifications to the basic training process above we can also train DNNs with ternary weights.

\subsection{Limiting the Weight Value Range with Parameterization}

\begin{wrapfigure}{r}{0.36\textwidth}
\vskip -0.5in
\centering
     \includegraphics[width=0.36\textwidth]{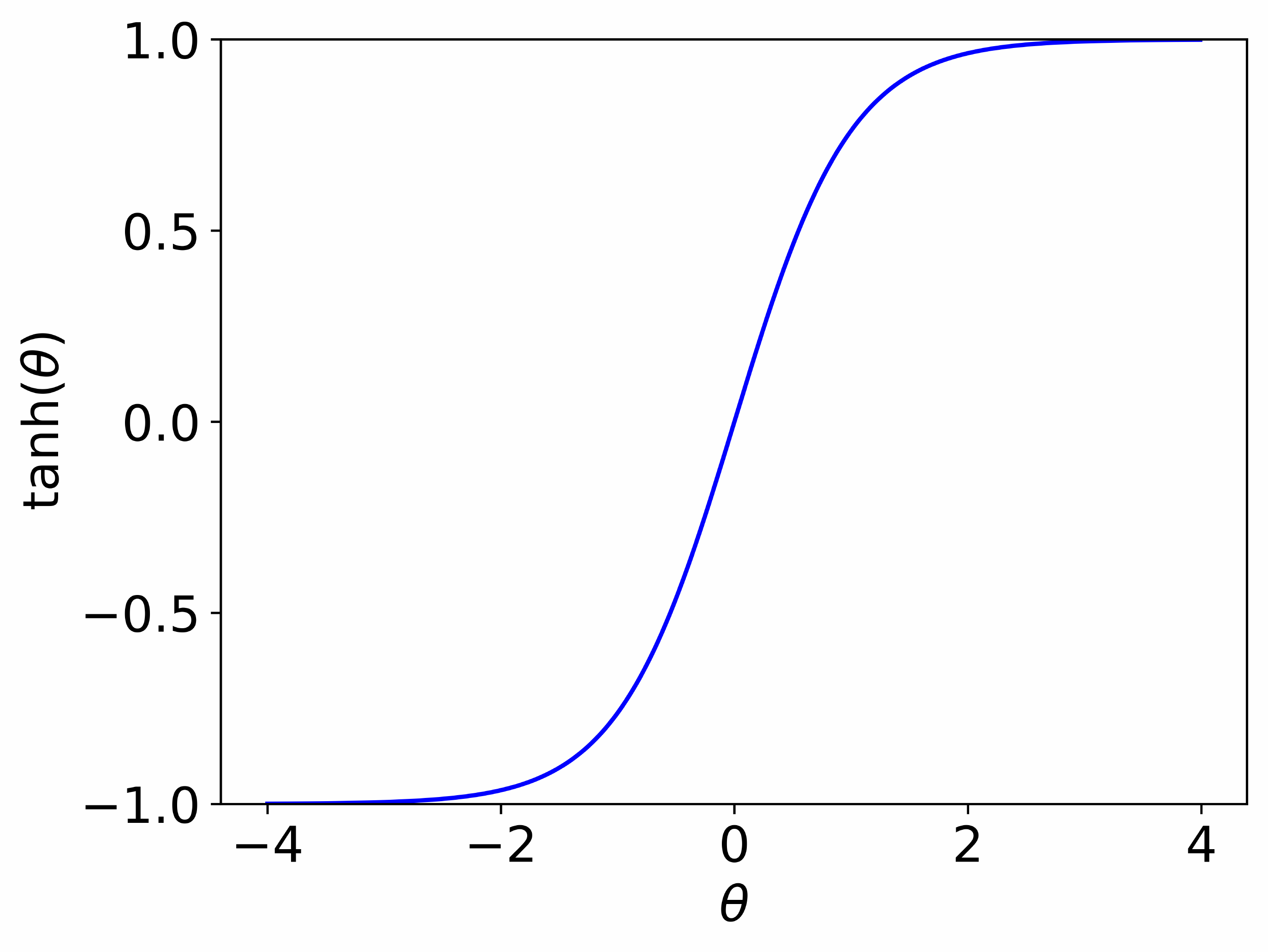}
     \vskip -0.1in
     \caption{Graph of  $\tanh(\theta)=\frac{\exp\left(\theta\right)-exp\left(-\theta\right)}{exp\left(\theta\right)+exp\left(-\theta\right)}$}
     \label{f1}
\end{wrapfigure}
DNNs with ternary weights are dramatically efficient at the test time as it needs less memory and no multiplication operation.
The challenge for training ternary weights is that the gradients with respect to the discrete weights are 0 or do not exist.
The existing studies utilize the straight-through estimator to solve this problem.
In this paper, we propose a novel method to use the real gradient instead of the estimator to do backpropagation.
As the ternary solution to a DNN is also in the full-precision weight space, we are inspired to maintain continuous value weights during training, enabling us to do normal backpropagation.
However, the full-precision weight space (usually 16 bits or 32 bits) is too large to find an appropriate ternary solution.
To alleviate this issue, we propose to compress the continuous weight space.
The possible values for ternary weights are -1, 0, and +1 that are between -1 and +1.
The range of the function $\tanh(.)$ is between -1 and +1 as shown in Figure \ref{f1}, so that we use $\tanh(\Theta)$ to parameterize the continuous weights $W$:
\begin{equation}
\label{2}
W = \tanh(\Theta)
\end{equation}
where $\Theta=[\theta_1, \theta_2, ..., \theta_n]$ are the parameters for computing $W$ and $\theta_i$ is used to compute $w_i$.\par

Then we replace $W$ in loss (\ref{1}) with $\tanh(\Theta)$, obtaining the following loss term:

\begin{equation}
\label{3}
L_c(\tanh(\Theta)) = \mathcal{L}(f(X, \tanh(\Theta)), Y)
\end{equation}
where $\tanh(\Theta)$ can be seen as the instantiations of $W$.
The advantage of doing this is that the weight space is significantly reduced and the backprobagation can still be applied to computing the gradients with respect to $\Theta$, enabling us to search for a ternary solution in a much smaller, continue weight space.
Note that although $\tanh(\Theta)$ function has been widely used in deep learning, how to use it to control the sparsity of ternary weights has never been studied.
We show this below.
\par

\subsection{Weight Discretization Regularization and Objective Function}
Although the weight space is extremely decreased by using parameterization $\tanh(\Theta)$, it is still difficult to find the discrete solution in this space.
If we directly solve for the minimization problem with (\ref{3}) as the objective, we are able to obtain a solution in the continuous space (-1, 1) that may still be far from the discrete solution.
To address this issue, we develop a novel weight discretization regularizer (WDR) to force the weights to be ternary and to control the sparsity of the ternary weights by taking advantage of the property of tanh() function:
\begin{equation}
\begin{aligned}
\label{4}
R(\tanh(\Theta))= \sum_{i=1}^n \sum_{j=1}^{|\theta_i|} \left[\left(\alpha-\tanh^2\left(\theta_{ij}\right)\right)\tanh^2\left(\theta_{ij}\right)\right]
 \end{aligned}
\end{equation}
where $|\theta_i|$ denotes the number of the elements in $\theta_i$; $\theta_{ij}$ represents the $j$th element in $\theta_i$; and $\alpha$ is the shape controller of function $R()$.
Note that one may have a \textbf{misunderstanding} at first glance that WDR enforces the weights to converge at \{$-\sqrt{\alpha}$, $+\sqrt{\alpha}$, 0\}.
We will show below (in Section \ref{wdr}) that this is \textbf{not true} and how WDR leads to the expected ternary weights \{-1, +1, 0\}.
More essentially, shape controller $\alpha$ is also the sparsity controller of the trained ternary weights (the sparsity here is measured by the percentage of 0s in the ternary weights).
We will theoretically (in Section \ref{wdr}) and empirically (in Section \ref{wdrexp1}) show that the sparsity of the ternary weights is positively related to $\alpha$.

We combine the loss term (\ref{3}) and WDR (\ref{4}) to obtain the final objective function of SCA:
\begin{equation}
\label{5}
J = L_c(\tanh(\Theta)) + \lambda R(\tanh(\Theta))
\end{equation}
where coefficient $\lambda$ is a hyperparameter to balance the contributions between the loss term $L_c(\tanh(\Theta))$ and the regularizer $R(\tanh(\Theta))$.
\par

\begin{figure}
\centering
     \includegraphics[width=1\textwidth]{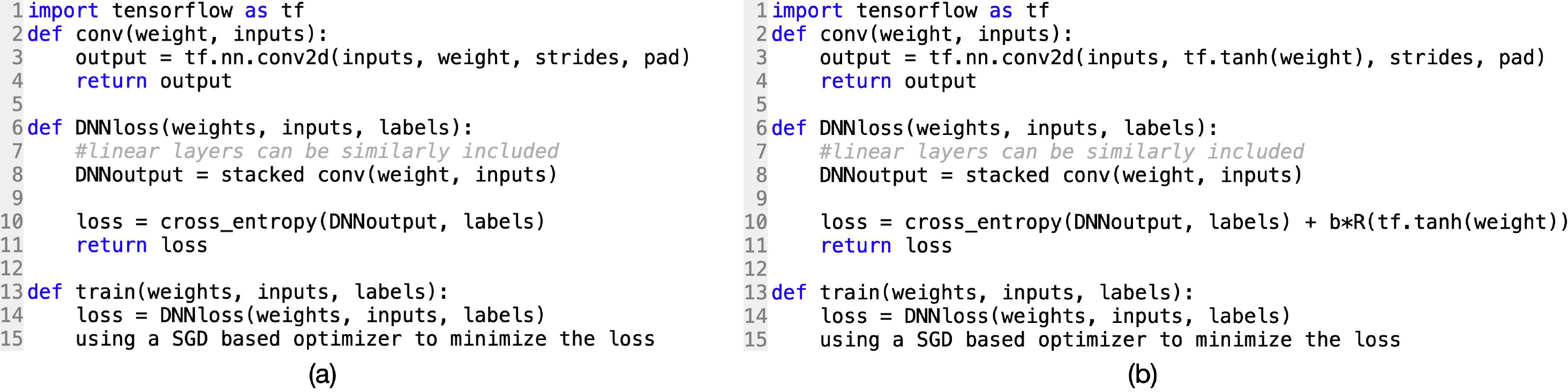}
     \vskip -0.1in
     \caption{Pseudocode compariosn. (a): Pseudocode for training full-precision networks based on tensorflow. (b): Pseudocode for training ternary weight networks via SCA based on tensorflow where $b$ in line 10 is the balancing weight and $R()$ is the proposed regularizer WDR that can be simply implemented as a function.}
    \label{code_com}
\end{figure}

At the training time, we solve for the minimization problem with objective (\ref{5}) using a gradient descent based optimizer as the gradients of $J$ with respect to $\Theta$ exist.
It is observed that SCA only needs two simple operations to convert a full-precision network to the ternary version, i.e., replacing weights $W$ with $\tanh(\Theta)$ and adding the WDR term $R(\tanh(\Theta))$.
Thus, it has an extremely high code reuse rate for converting full-precision networks to the ternary counterparts.
To further illustrate this point, we compare the pseudocode for training full-precision weight networks with that of SCA.
As shown in Figure \ref{code_com}, their differences lie in line 3 and line 10.
In line 3, SCA simply modifies the conv function by adding $\tanh()$ to parametrize the weights.
In line 10, SCA simply adds a regularizer to force the weights to be ternary.
SCA only changes the conv function and then adds a regularizer to the loss.
Therefore, it has a very high code reuse rate when converting a full-precision network to the ternary counterpart.

\par

\subsection{Test-Time Inference}
After (\ref{5}) converges, we take the learned $\Theta$ to compute the ternary weights for inference.
Specifically, at the test time, we obtain the ternary weights $W_{ter}$ by rounding $\tanh(\Theta)$ to the nearest integer:
\begin{equation}
\label{6}
W_{ter}=round(\tanh(\Theta))
\end{equation}
Since the range of $\tanh(\theta_{ij})$ is between -1 and +1, the obtained nearest integer of $\tanh(\theta_{ij})$ can be -1, 0, or +1.
For given test data $x_t$, the prediction is obtained by using $W_{ter}$:
\begin{equation}
\label{7}
        Pred=f(W_{ter}, x_t)
\end{equation}
Thanks to the regularizer WDR, simply rounding $\tanh(\Theta)$ to the nearest integer almost does not lead to any performance dropping.
We summarize SCA in Algorithm 1.

\begin{algorithm} 
\caption{SCA} 
 \begin{algorithmic} [1] 
 \renewcommand{\algorithmicrequire}{\textbf{Input:}} 
\renewcommand{\algorithmicensure}{\textbf{Output:}}
\REQUIRE Training data (X, Y), a neural network $f$ with initial parameters $\Theta_0$
\ENSURE Ternary weights $W_{ter}$
\STATE Construct objective function (\ref{5}) and minimize it with gradient descent to obtain $\Theta_{opt}$ 
\STATE Obtain ternary weights $W_{ter}$ by simply rounding $\tanh(\Theta_{opt})$ to nearest integers
\end{algorithmic} 
\end{algorithm}

\begin{figure}[]
     \begin{minipage}{0.34\textwidth}
     \centering
     \includegraphics[height=4cm]{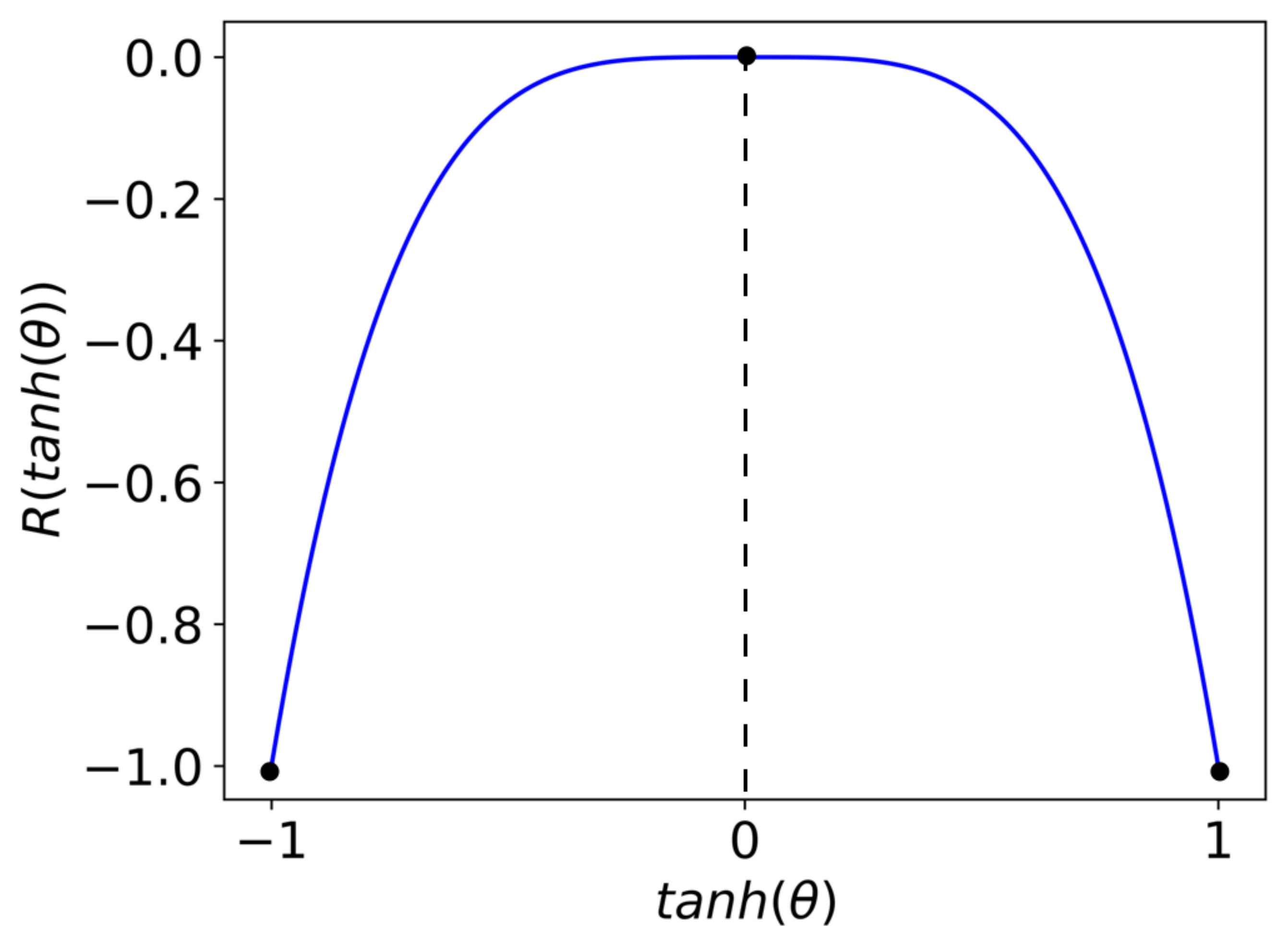}
     \caption{Function graph of $R(\tanh(\theta))$ \newline with $\alpha=0$}
     \label{f2}
   \end{minipage}\hfill
   \begin{minipage}{0.32\textwidth}
     \centering
     \includegraphics[height=4cm]{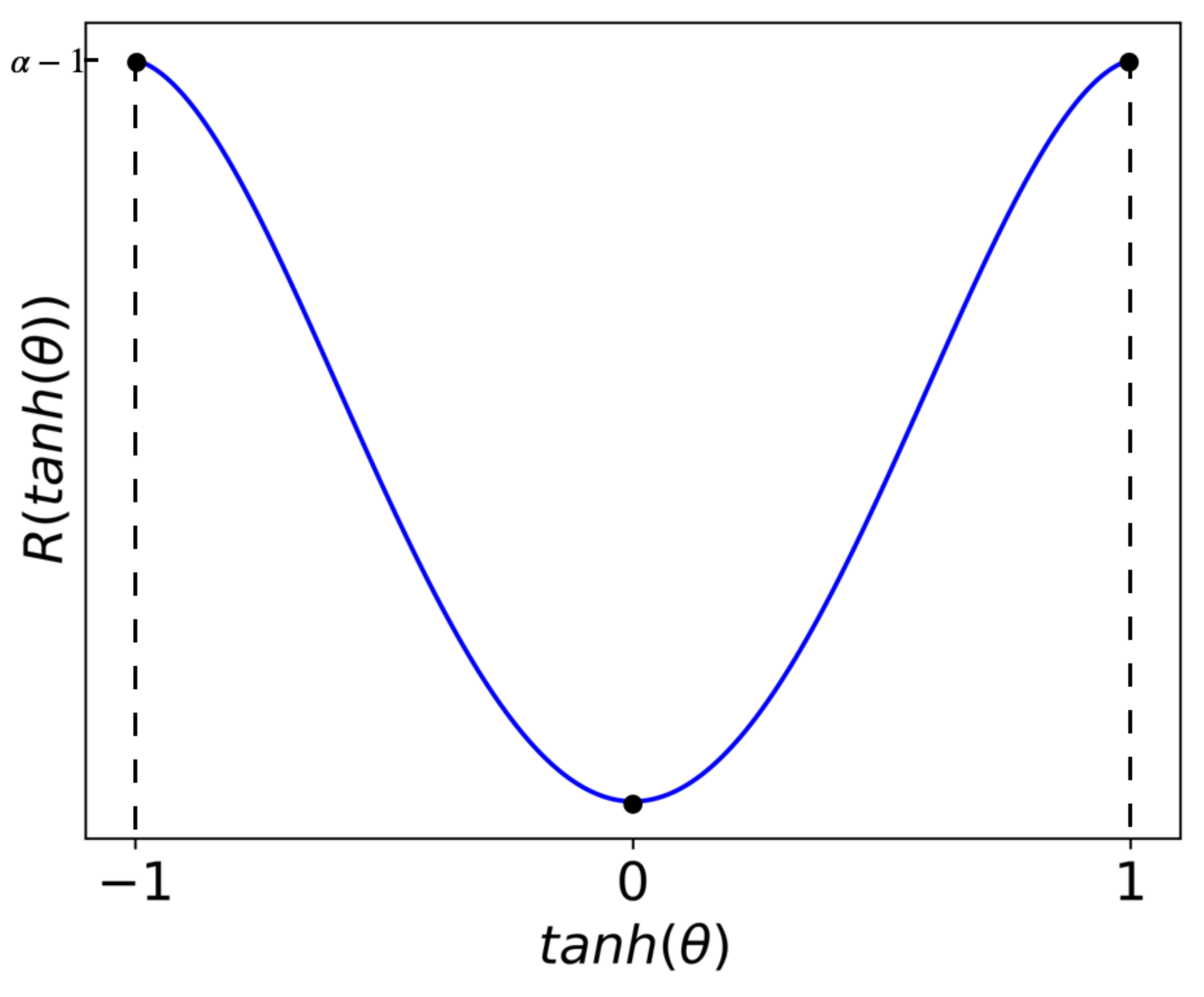}
     \caption{Function graph of $R(\tanh(\theta))$\newline with $\alpha\geq2$}
     \label{f3}
   \end{minipage}\hfill
   \begin{minipage}{0.32\textwidth}
     \centering
     \includegraphics[height=4cm]{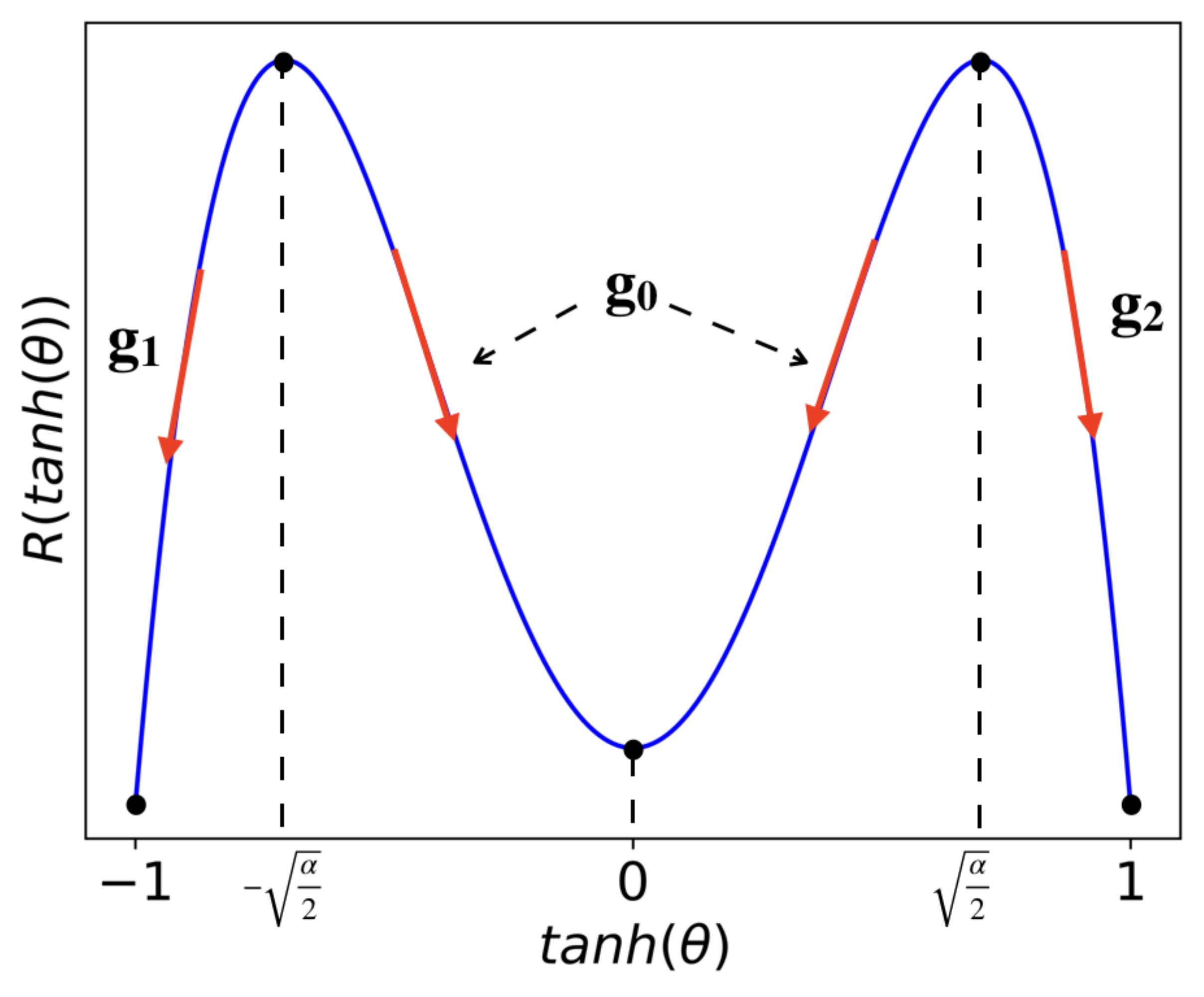}
     \caption{Function graph of $R(\tanh(\theta))$ \newline with $\alpha$ in (0, 2)}
     \label{f4}
   \end{minipage}  
\end{figure}

\subsection{Theoretical Analysis}
\label{wdr}

In this part, we further examine the proposed regularizer WDR and theoretically show why $\alpha$ is able to control the sparsity of the trained ternary weights.
For clarity, we omit the subscripts $i$, $j$ of $\theta_{ij}$ and observe WDR w.r.t. each $\theta_{ij}$:
\begin{equation}
\label{8}
R(\tanh(\theta))=\left( \alpha- \tanh^2 \left( \theta \right) \right)\tanh^2\left(\theta\right)
\end{equation}
The derivative function of (\ref{8}) is expressed as:
\begin{equation}
\begin{aligned}
\label{9}
\frac{\partial R(\tanh(\theta))}{\partial \theta}=2\tanh\left(\theta\right)\left(1-\tanh^2\left(\theta\right)\right)\left(\alpha-2\tanh^2\left(\theta\right)\right)
\end{aligned}
\end{equation}
For training ternary weights \{-1, 0, +1\}, it is desired that $\tanh(\theta)=0$, $\tanh(\theta)=-1$ \footnote{Note that $\tanh(\theta)$ cannot equal to -1 or 1, but can be infinitely close to -1 or 1}, and $\tanh(\theta)=+1$ are three minimum points of the WDR term $R(\tanh(\theta))$.
We provide below the minimum and maximum points of $R(\tanh(\theta))$ in different cases, i.e., $\alpha=0$, $0< \alpha < 2$, and $\alpha \geq 2$

\begin{proposition}
\label{t1}
When $\sqrt{\frac{\alpha}{2}}=0$ (i.e., $\alpha=0$), $R(\tanh(\theta))$ has three zero-gradient points with $\tanh(\theta)=0$ as the maximum point, and $\tanh(\theta)=-1$ and $\tanh(\theta)+1$ as the two minimum points.
\end{proposition}

{\em Proof of of proposition \ref{t1}}: When $\alpha=0$, the derivative function of $R(\tanh(\theta))$ is written as:
\begin{equation}
\begin{aligned}
\label{11}
\frac{\partial R(\tanh(\theta))}{\partial \theta}=-4\tanh^3\left(\theta\right)\left(1-\tanh^2\left(\theta\right)\right)
\end{aligned}
\end{equation}
By solving $\frac{\partial R(\tanh(\theta))}{\partial \theta}=0$, we obtain three zero-gradient points, i.e., $\tanh(\theta)$ = 0, $\tanh(\theta)$ = -1, and $\tanh(\theta)$ = +1.
In addition, $\left(1-\tanh^2\left(\theta\right)\right)$ in right hand side of (\ref{11}) is always positive as the range of $\tanh(\theta)$ is (-1, 1) as shown in Figure \ref{f1}.
Thus, the sign of (\ref{11}) is determined by $-4\tanh^3\left(\theta\right)$.
When $\tanh(\theta)\in(-1, 0)$, $-4\tanh^3\left(\theta\right)$ is positive so that (\ref{11}) is positive.
It means that $R(\tanh(\theta))$ is increasing when $\tanh(\theta)\in(-1, 0)$.
Similarly, When $\tanh(\theta)\in(0, 1)$, $-4\tanh^3\left(\theta\right)$ is negative so that (\ref{11}) is negative.
It means that $R(\tanh(\theta))$ is decreasing when $\tanh(\theta)\in(0, 1)$.
We provide the function graph of $R(\tanh(\theta))$ with $\alpha=0$ in Figure \ref{f2}.
Therefore, $\tanh(\theta)=0$ is the maximum point and $\tanh(\theta)=-1$ and $\tanh(\theta)=+1$ are two minimum points.
The above proves Proposition \ref{t1}.
\par

However, the case in Proposition \ref{t1} is not desired for training ternary weights as the ideal case for training ternary weights is that $\tanh(\theta)=0$, $\tanh(\theta)=-1$, and $\tanh(\theta)=+1$ are three minimum points.

\begin{proposition}
\label{t2}
When $\sqrt{\frac{\alpha}{2}}\geq1$ (i.e., $\alpha\geq2$), $R(\tanh(\theta))$ has three zero-gradient points with $\tanh(\theta)$=0 as the minimum point, and $\tanh(\theta)$=-1 and $\tanh(\theta)$=+1 as the two maximum points
\end{proposition}

{\em Proof of Proposition \ref{t2}}: When $\sqrt{\frac{\alpha}{2}}\geq1$ (i.e., $\alpha\geq2$), by directly solving (\ref{9}) = 0, we obtain five zero-gradient points of $R(\tanh(\theta))$, i.e., $\tanh(\theta)$ = 0, $\tanh(\theta)$ = -1, $\tanh(\theta)$ = +1, $\tanh(\theta)$ = $-\sqrt{\frac{\alpha}{2}}$, and $\tanh(\theta)$ = $\sqrt{\frac{\alpha}{2}}$.
However, as $\sqrt{\frac{\alpha}{2}}\geq1$ but the range of $\tanh(\theta)$ is (-1, 1), zero-gradient points $\tanh(\theta)$ = $-\sqrt{\frac{\alpha}{2}}$ and $\tanh(\theta)$ = $\sqrt{\frac{\alpha}{2}}$ do not exist.
Thus, in this case, $\tanh(\theta)$ = 0, $\tanh(\theta)$ = -1, and $\tanh(\theta)$ = +1 are three zero-gradient points of $R(\tanh(\theta))$.
In addition, $\left(1-\tanh^2\left(\theta\right)\right)\left(\alpha-2\tanh^2\left(\theta\right)\right)$ in (\ref{9}) is always positive when $\sqrt{\frac{\alpha}{2}}\geq1$ (i.e., $\alpha\geq2$).
Thus, the sign of (\ref{9}) is determined by $2\tanh(\theta)$.
When $\tanh(\theta)\in(-1, 0)$, $2\tanh(\theta)$ is negative so that (\ref{9}) is negative.
It means that $R(\tanh(\theta))$ is decreasing when $\tanh(\theta)\in(-1, 0)$.
Similarly, when $\tanh(\theta)\in(0, 1)$, $2\tanh(\theta)$ is positive so that (\ref{9}) is positive.
It means that $R(\tanh(\theta))$ is increasing when $\tanh(\theta)\in(0, 1)$.
We provide the function graph of $R(\tanh(\theta))$ with $\alpha \geq2$ in Figure \ref{f3}.
Therefore, $\tanh(\theta)=0$ is the minimum point and $\tanh(\theta)=-1$ and $\tanh(\theta)=+1$ are two maximum  points.
The above proves Proposition \ref{t2}.\par

Obviously the case in Proposition \ref{t2} is not desired for training ternary weights, either.
Let us pay more attention to the last case.

\begin{proposition}
\label{t3}
When $0<\sqrt{\frac{\alpha}{2}}<1$ (i.e, $0<\alpha<2$), $R(\tanh(\theta))$ has five zero-gradient points.
$\tanh(\theta)=0$, $\tanh(\theta)=-1$, and $\tanh(\theta)=+1$ are three minimum points, and $\tanh(\theta)=-\sqrt{\frac{\alpha}{2}}$ and $\tanh(\theta)=\sqrt{\frac{\alpha}{2}}$ are two maximum points.
Moreover, the percentage of 0s in the resulting ternary weights by minimizing $R(\tanh(\theta))$ is positively related to $\alpha$.
\end{proposition}

{\em Proof of proposition \ref{t3}}: When $0<\sqrt{\frac{\alpha}{2}}<1$, by directly solving (\ref{9}) = 0, we obtain five zero-gradient points for $R(\tanh(\theta))$, i.e., $\tanh(\theta)$ = 0, $\tanh(\theta)$ = -1, $\tanh(\theta)$ = +1, $\tanh(\theta)$ = $-\sqrt{\frac{\alpha}{2}}$, and $\tanh(\theta)$ = $\sqrt{\frac{\alpha}{2}}$, and all the five zero-gradient points are meaningful.
In addition, as $\left(1-\tanh^2\left(\theta\right)\right)$ in (\ref{9}) is always positive, the sign of (\ref{9}) is determined by $2\tanh(\theta)\left(\alpha-2\tanh^2\left(\theta\right)\right)$.
When $\tanh(\theta)\in(-1, -\sqrt{\frac{\alpha}{2}})$, $2\tanh(\theta)\left(\alpha-2\tanh^2\left(\theta\right)\right)$ is positive so that (\ref{9}) is positive.
It means that $R(\tanh(\theta))$ is increasing when $\tanh(\theta)\in(-1, -\sqrt{\frac{\alpha}{2}})$.
When $\tanh(\theta)\in(-\sqrt{\frac{\alpha}{2}}, 0)$, $2\tanh(\theta)\left(\alpha-2\tanh^2\left(\theta\right)\right)$ is negative so that (\ref{9}) is negative.
It means that $R(\tanh(\theta))$ is decreasing when $\tanh(\theta)\in(-\sqrt{\frac{\alpha}{2}}, 0)$.
When $\tanh(\theta)\in(0, \sqrt{\frac{\alpha}{2}})$, $2\tanh(\theta)\left(\alpha-2\tanh^2\left(\theta\right)\right)$ is positive so that (\ref{9}) is positive.
It means that $R(\tanh(\theta))$ is increasing when $\tanh(\theta)\in \left(0, \sqrt{\frac{\alpha}{2}}\right)$.
When $\theta\in(\sqrt{\frac{\alpha}{2}}, 1)$, $2\tanh(\theta)\left(\alpha-2\tanh^2\left(\theta\right)\right)$ is negative so that (\ref{9}) is negative.
It means that $R(\tanh(\theta))$ is decreasing when $\tanh(\theta)\in(\sqrt{\frac{\alpha}{2}}, 1)$.
We provide the function graph of $R(\tanh(\theta))$ with $0<\sqrt{\frac{\alpha}{2}}<1$ (i.e, $0<\alpha<2$) in Figure \ref{f4}.
Therefore, $\tanh(\theta)=0$, $\tanh(\theta)=-1$, and $\tanh(\theta)=+1$ are three minimum points, and $\tanh(\theta)=-\sqrt{\frac{\alpha}{2}}$ and $\tanh(\theta)=\sqrt{\frac{\alpha}{2}}$ are two maximum points.
Moreover, as shown in Figure \ref{f4}, the points with $\tanh(\theta)$ in $\left(-\sqrt{\frac{\alpha}{2}}, \sqrt{\frac{\alpha}{2}}\right)$ are more inclined to moving toward the minimum point with $\tanh(\theta)=0$ due to gradient $g_0$. The points with $\tanh(\theta)$ in $\left(-1, -\sqrt{\frac{\alpha}{2}}\right)$ are more inclined to moving toward the minimum point with $\tanh(\theta)=-1$ due to gradient $g_1$, and the points with $\tanh(\theta)$ in $\left(\sqrt{\frac{\alpha}{2}}, 1\right)$ are more inclined to moving toward the minimum point with $\tanh(\theta)=1$ due to gradient $g_2$.
Consequently, the percentage of 0s in the resulting ternary weights is positively related to the length of the range $\left(-\sqrt{\frac{\alpha}{2}}, \sqrt{\frac{\alpha}{2}}\right)$, i.e, $\sqrt{2\alpha}$, and it is thus positively related to $\alpha$.
The above proves Proposition \ref{t3}.\par

It is obvious that the case in Proposition \ref{t3} is desired for training ternary weight networks as $\tanh(\theta)=-1$, $\tanh(\theta)=0$, and $\tanh(\theta)=+1$ are three minimum points.
We also empirically verify the positive correlation between the sparsity of the ternary weights and the values of $\alpha$ in the experiment section.

\subsection{Differences between WDR in SCA and the Existing Regularizers}
\label{difs}
In this part, we provide the differences between regularizer WDR in SCA with the regularizers in the existing literature.\par 

Tang et al. \cite{tang2017train} proposed a regularizer $1-W^2$ for training binary weight networks, which can be easily extended to $(1-W^2)W^2$ for training ternary weight networks.
However, this simple regularizer cannot control the sparsity of the ternary weights and it relies on the straight-through estimator to estimate the gradients with respect to discrete weights $W$.
In contrast, WDR addresses these two limitations by introducing the combination of a controller $\alpha$ and parameterization $\tanh(\Theta)$ to control the sparsity of the ternary weights and allow for using the real gradients.
\par

BNN+ \cite{darabi2018bnn+} introduced two regularizers for training binary neural networks, i.e., $|\alpha-|W||$ and $(\alpha-|W|)^2$, which can be easily extended to $|\alpha-|W||*|W|$ and $(\alpha-|W|)^2*|W|^2$, respectively, for training ternary weight networks.
First, it is obvious that the straight-through estimator is still required to estimate the gradients with respect to discrete weights $W$.
The proposed WDR uses the real gradients by introducing $\tanh(\Theta)$.
Second, besides $|W|=0$, the minimum point for these two regularizers in BNN+ is $|W|=\alpha$, which leads to ternary weights \{$-\alpha, 0, +\alpha$\}.
Consequently, BNN+ introduces $\alpha$ to \textbf{control the scaling} of $W$.
In contrast, $\alpha$ in our WDR is used to \textbf{control the sparsity} of the ternary weights, which results in more efficient ternary weight networks.
Moreover, \{$-\alpha, 0, +\alpha$\} in BNN+ is less efficient than \{-1, 0, +1\} in WDR.
\par

\section{Experiments}
In this section, we report extensive experiments for evaluating SCA on several benchmark datasets.
Through these experiments, we aim to address the following research questions:
\begin{itemize}
\item How much is the performance gap between a ternary weight network trained by SCA and the full-precision weight counterpart?
\item Does our approach SCA yield better performances than those of the existing approaches?
\item Is the sparsity of the trained ternary weights positively related to $\alpha$ on real datasets?
\item What is the difference between the features learned by SCA and the features of full-precision networks?
\end{itemize}

\subsection{Datasets}
We adopt five benchmark datasets: MNIST \cite{lecun-mnisthandwrittendigit-2010}, CIFAR-10 \cite{krizhevsky2009learning}, CIFAR-100 
\cite{krizhevsky2009learning}, Tiny ImageNet \footnote{http://tiny-imagenet.herokuapp.com/}, and ImageNet (ILSVRC2012) \cite{deng2009imagenet}.

\textbf{MNIST} is a handwritten digit image classification dataset with 10 classes, containing 60,000 training images and 10,000 test images.
We do not use any data augmentation on MNIST.\par

$\textbf{CIFAR-10}$ is an image classification dataset with 10 classes, containing 50,000 training images and 10,000 test images with image size 32 $\times$ 32 in the RGB space.
We follow the standard data augmentation on CIFAR-10.
During training time, we pad 4 pixels on each side of an image and randomly flip it horizontally.
Then the image is randomly cropped to 32 $\times$ 32 size. 
During test time, we only evaluate the single view of an original 32 $\times$ 32 image without padding or cropping.

$\textbf{CIFAR-100}$ comprises similar images to those in CIFAR-10, but has 100 classes.
We adopt the same data augmentation strategy as that in CIFAR-10.

$\textbf{Tiny ImageNet}$, i.e., a subset of ImageNet, is an image classification dataset with 200 classes, containing 100,000 training images and 10,000 test images with size 64 $\times$ 64 in the RGB space.
We adopt the standard data augmentation strategy on Tiny ImageNet, i.e., randomly padding, flipping, and cropping.
At test time, we only evaluate the original image.

$\textbf{ImageNet}$ is a large-scale image classification dataset with 1000 classes, containing 1.28 million training images and 50,000 validation images with different sizes in the RGB space.
On ImageNet, we use the standard scale and aspect ratio augmentation strategy from \cite{szegedy2015going}.
Test images are resized so that the shorter side is set to 256, and then are cropped to size 224 $\times$ 224.\par

\par

\subsection{Competitors}

For fair comparison, we only compare SCA with the existing approaches\footnote{We only compare SCA with the competitors in term of test accuracy because almost all the competitors except TTQ \cite{zhu2016trained} only report the accuracy without sparsity.} for training DNNs with binary or ternary weights without discrete activations. 
The competitors for training DNNs with binary weights include BinaryConnect \cite{courbariaux2015binaryconnect}, BWN \cite{rastegari2016xnor}, DoReFa \cite{zhou2016dorefa}, and BayesBiNN \cite{meng2020training}. The competitors for training DNNs with ternary weights include TWN \cite{li2016ternary}, TTQ \cite{zhu2016trained}, INQ \cite{zhou2017incremental}, LR-Net \cite{shayer2017learning}, ELQ \cite{zhou2018explicit}, and ELB \cite{leng2018extremely}.
As SCA is a regularizer-based approache, we also compare SCA with the regularizers in \cite{tang2017train, darabi2018bnn+} which are extended to training ternary weight networks.
The performances of all the competitors are all taken from their original papers or obtained from the author-released codes unless otherwise specified.
We follow the same convention as that used in the existing work to keep the first and last layers in full precision.
All the results below are reported based on 3 runs.
\par

\subsection{Comparison with State-of-the-art Approaches}
We report the performances of different approaches as well as the full-precision networks on different datasets.


\begin{table}
\centering
\caption{Test accuracies (\%) on MNIST}
\label{m1}
\begin{tabular}{lcc}
\toprule
Methods        & Weight Types                         & Accuracy (\%) \\ \midrule
BinaryConnect  &\multirow{2}{*}{Binary: \{-1 ,+1\}  }

   & 98.71         \\ 
BayesBiNN & & 98.86        \\ 
\midrule
BWN            &Ternary: \{-$\alpha$, +$\alpha$\}                     & 99.05         \\ \midrule

Extended \cite{tang2017train}  &\multirow{3}{*}{Ternary: \{-1, 0, +1\}} & 99.11         \\
LR-Net         & & 99.50         \\
SCA (Ours)    &                                & \textbf{99.56$\pm$0.02}         \\ \midrule
TWN            &\multirow{2}{*}{Ternary: \{-$\alpha$, 0, +$\alpha$\}}             & 99.35         \\
Extended \cite{darabi2018bnn+}  & & 99.15         \\ \midrule
Full-precision network & Full-Precision                 & 99.56        \\ \bottomrule
\end{tabular}
\end{table}




\subsubsection{Performances on MNIST }

On MNIST, we use the same architecture as that in the competitors \cite{li2016ternary, shayer2017learning}, i.e., $(32-C5) + MP2 + (64-C5) + MP2 + 512FC + Softmax$, where ($32-C5$) is the convolutional layer containing 32 filters of size $5\times5$; $MP2$ is the max-pooling layer with stride 2; and $512FC$ denotes the fully connected layer with 512 nodes.
We adopt dropout \cite{JMLR:v15:srivastava14a} before the last layer with drop rate of 0.5.
$\lambda$ and $\alpha$ are set to 1e-7 and 1e-4, respectively.
The weights are initialized with Xavier initializer \cite{glorot2010understanding}.
The objective function is minimized with optimizer Adam \cite{kingma2014adam} and mini-batch size 128.
The initial learning rate is 0.01 and is divided by 10 at the 100th epoch and the 160th epoch.
We have trained the network for 200 epochs.
\par

The comparison results on MNIST are reported in Table \ref{m1}.
It is observed that SCA achieves the best performance among the approaches with ternary weights \{-1, 0 ,+1\}, also beats the other state-of-the-art approaches significantly, and even matches the performance of the full-precision weight counterpart.
This demonstrates that SCA is able to compress a full-precision weight network to the ternary counterpart without accuracy dropping.
We also notice that the performance gap between SCA (99.56\%) and the best competitor (99.50\%) is not large.
The reason is that the performances of SCA and the best competitor are both very close to that of the full-precision network (99.56\%).
The performance of the full-precision network can be considered as the upper bound of the ternary weight network performances.
Both SCA and the best competitor can match this upper bound so that their performance gap is not large.
However, SCA still shows its superiority over the existing approaches as it achieves the same accuracy as that of the full-precision network.


\subsubsection{Performances on CIFAR-10}

On CIFAR-10, we adopt the same architectures as those in the competitors \cite{li2016ternary, shayer2017learning, zhu2016trained}: VGG-S \cite{li2016ternary, shayer2017learning}, ResNet-20 \cite{zhu2016trained, he2016deep}, and VGG-Variate \cite{cai2017deep}.
For VGG-S, $\lambda$ and $\alpha$ are 5e-8 and 0.1, respectively; dropout with drop rate of 0.5 is adopted; weight decay is used in the last layer with parameter 1e-5; the objective function is minimized with Adam with mini-batch size 128; the initial learning rate is 0.01 and is divided by 10 at the 200th epoch and the 370th epoch; we have trained the network for 450 epochs.
For ResNet-20, $\lambda$ and $\alpha$ are set to 5e-6 and 0.1, respectively; weight decay is used in the last layer with parameter 1e-5; the initial learning rate is 0.005 and is divided by 5 at the 150th epoch, then divided by 2 at the 450th epoch; we have trained the network for 700 epochs with mini-batch size 128.
For VGG-Variate, $\alpha$ is set to 0.1; dropout \cite{JMLR:v15:srivastava14a} with drop rate of 0.5 is adopted before the last layer; the initial $\lambda$ is set to 5e-8 and is multiplied by 50 after 270 epochs of training; weight decay is used in the last layer with parameter 1e-5; the objective function is minimized with Adam with mini-batch size 128; the initial learning rate is 0.005 and is divided by 10 at the 120th epoch and the 270th epoch; we have trained the network for 370 epochs.
We adopt the same initialization strategy as that in LR-Net \cite{shayer2017learning} and TTQ \cite{zhu2016trained} by using the pretrained full-precision weights as the initialization for VGG-Variate and all ResNets while VGG-S is initialized with Xavier initializer \cite{glorot2010understanding}.

\par

\begin{table*}[!t]
\centering
\setlength{\abovecaptionskip}{0.1cm}
\setlength{\belowcaptionskip}{0.1cm}
\caption{Test accuracies (\%) on CIFAR-10}
\label{m2}
\begin{tabular}{lcccc}
\toprule
Methods        & Weight Types                         & VGG-S         & ResNet-20     &VGG-Variate  \\ \midrule
BinaryConnect  &Binary: \{-1 ,+1\}                     & 91.10          & -         & -      \\ \midrule
BWN            & \multirow{2}{*}{Binary: \{$\alpha$, +$\alpha$\}}    & 90.18          & -     & -   \\ 
DoReFa        &                      &-                           &90.00 & -  \\ \midrule
Extended \cite{tang2017train}  &\multirow{3}{*}{Ternary: \{-1, 0, +1\}} & 90.17 & 89.97 & 91.06         \\
LR-Net         &  & 93.26          & 90.08       & 91.47      \\
SCA (Ours)    &                                & \textbf{93.41$\pm$0.10} & \textbf{91.28 $\pm$ 0.15} & \textbf{92.75 $\pm$ 0.17} \\ \midrule
TWN            &\multirow{2}{*}{Ternary: \{-$\alpha$, 0, +$\alpha$\} } & 92.56          & -      & -        \\ 
Extended \cite{darabi2018bnn+}  &  & 90.32 & 90.10 & 91.13          \\ \midrule
TTQ            &Ternary: \{-$\alpha$, 0, +$\beta$\}                  & -              & 91.13     & -     \\ \midrule
 Full-precision network & full-precision                 & 93.42          & 91.76    & 92.75     \\ \bottomrule
\end{tabular}
\end{table*}

Table \ref{m2} summarizes the comparison results on CIFAR-10.
We observe that SCA compresses the full-precision networks to the ternary versions almost without accuracy dropping for VGG-S and VGG-Variate but with a little accuracy dropping for ResNet-20.
The reason is that VGG-S and VGG-Variate ($\geq$ 0.84M parameters) have much more parameters than those in ResNet-20 (0.27M parameters) and compressing over-parameterized networks may not hurt their performances.
On the other other hand, despite its simplicity, SCA also significantly outperforms all the binary or ternary competitors on all the three networks, which demonstrates the effectiveness and superiority of SCA.


\begin{table*}[t]
\centering
\caption{Performances on ImageNet}
\label{m3}
\begin{tabular}{lccc|cc}
\toprule
    &                          & \multicolumn{2}{c|}{ResNet-18}            & \multicolumn{2}{c}{AlexNet}                       \\
Methods  & Weight Types       & TOP-1 (\%)      & TOP-5 (\%)         & TOP-1 (\%)    & TOP-5 (\%)                \\ \midrule
BWN   &       Binary: \{-$\alpha$, +$\alpha$\}  & 60.8                         & 83.0      &  56.8                    & 79.4            \\ \midrule
Extended \cite{tang2017train}  & \multirow{3}{*}{\begin{tabular}[c]{@{}c@{}}Ternary: \{-1, 0, +1\} \end{tabular}} & 65.8    &  86.7  & 54.8    &  76.9 \\
LR-Net  &  &63.5          &    84.8   & 55.9            & 76.3                 \\
SCA (Ours)             &             &  \textbf{67.9$\pm$0.11}     &          \textbf{88.0$\pm$0.21}     &   \textbf{59.3$\pm$0.16}   &        \textbf{80.8$\pm$0.19}      \\ \midrule
TWN    & \multirow{4}{*}{\begin{tabular}[c]{@{}c@{}}Ternary: \{-$\alpha$, 0, +$\alpha$\} \end{tabular}}  & 65.3       & 86.2  &  54.5         & 76.8     \\
ELB  &   & 67.0                       &87.5    &     58.2           &   80.6                    \\
ELQ     &      & 67.3                      & \textbf{88.0}                 & 57.9                 &   80.2           \\              
Extended \cite{darabi2018bnn+}  & & 66.0    &  86.3  & 54.9    &  76.6     
\\ \midrule
TTQ & \multirow{2}{*}{\begin{tabular}[c]{@{}c@{}}Ternary: \{-$\alpha$, 0, +$\beta$\} \end{tabular}}  & 66.6               & 87.2           &  57.5        & 79.7          \\
INQ     &  & 66.0                            & 87.1               &         -          &   -                            \\ \midrule
\multicolumn{1}{c}{Full-precision network} & \multicolumn{1}{c}{Full-Precision}      & \multicolumn{1}{c}{69.5} & \multicolumn{1}{c|}{89.2} & \multicolumn{1}{c}{60.8} & \multicolumn{1}{c}{81.9} \\ \bottomrule
\end{tabular}
\end{table*}


\subsubsection{Performances on ImageNet}
To investigate the performances of SCA on large-scale datasets, we conduct a series of experiments on ImageNet.
Limited by computation resources, we only adopt two networks on ImageNet.
We follow the competitors \cite{zhu2016trained,li2016ternary} and use 18-layer ResNet \cite{he2016deep} and AlexNet \cite{krizhevsky2012imagenet} architectures.
We adopt the same initialization strategy as that in TTQ \cite{zhu2016trained} and LR-Net \cite{shayer2017learning} by using the full-precision weights as the initialization.
For ResNet-18, $\lambda$ is initially set to 1e-9 and is multiplied by 1e2 and 10 at the 50th epoch and the 70th epoch, respectively; $\alpha$ is set to 0.3; weight decay is used in the last layer with parameter 1e-6; the objective function is minimized with Adam with mini-batch size 128; the initial learning rate is 0.005 and is divided by 10 after the 30th epoch, the 50th epoch, and the 70th epoch; we have trained the network for 90 epochs. 
For AlexNet, $\lambda$ is initially set to 1e-9 and is multiplied by 1e4 and 1e3 at the 70th epoch and the 100th epoch, respectively; $\alpha$ is set to 1.0; weight decay is used in the last layer with parameter $1e-6$; the objective function is minimized with Adam; the initial learning rate is 0.005 and is divided by 2, 10, and 5 at the 50th epoch, the 70th epoch, and the 100th epoch, respectively; we have trained the network for 130 epochs.\par

Table \ref{m3} reports the comparison results \footnote{The results of LR-Net on AlexNet and Extended \cite{tang2017train, darabi2018bnn+} are obtained from our implementation based on the paper.
} on ImageNet.
Clearly, SCA outperforms the competitors with ternary weights \{-1, 0, +1\} in terms of both Top1 and Top5 accuracies by a large margin, and also beats the competitors with the other kinds of ternary weights \{$\alpha$, 0 $-\alpha$/$\beta$\} significantly, which demonstrates the usefulness and applicability of SCA on large-scale datasets.
We also notice that even on large-scale datae ImageNet, the performances of the ternary weight networks trained by SCA are still close to those of the full-precision counterparts, which demonstrates the effectiveness of SCA for converting full-precision networks to the ternary versions on large-scale datasets.

\subsubsection{Performances on CIFAR-100 and Tiny ImageNet}
To further explore the performances of SCA, we evaluate it on more datasets, i.e., CIFAR-100 and Tiny ImageNet.
As the existing approaches did not report the results on these two datasets, we only compare the performances of SCA with the upper bound of ternary weight network performances, i.e., the performances of full-precision networks.
If SCA can match the upper bound, it means that SCA at least does not perform worse than the existing approaches.
On CIFAR-100, the values of the hyperparameters are the same as those on CIFAR-10 except that $\alpha$ is set to 1e-4 and 0.5 for VGG-S and VGG-variate, respectively.
On Tiny ImageNet, $\alpha$ is set to 0.1; $\lambda$ is initially set to 5e-7 and is multiplied by 1e2 at the 90th epoch; weight decay is used with parameter 1e-5; dropout with drop rate 0.5 is adopted; we have the networks for 120 epochs.
\par

Table \ref{mc100} reports the comparison results on CIFAR-100 and Tiny ImageNet.
It is observed that the ternary weight networks trained by SCA consistently match the performances of the full-precision counterparts on both datasets and both networks.
This demonstrates the effectiveness of SCA for compressing full-precision weight networks to the ternary weight counterparts.

\begin{table*}[!t]
 \begin{minipage}[t]{0.55\textwidth}
  \centering
     \makeatletter\def\@captype{table}\makeatother\caption{Test accuracies (\%) on CIFAR-100 and Tiny ImageNet}
     \label{mc100}
     \resizebox{0.99\textwidth}{!}{%
\begin{tabular}{llcc}
\toprule
                               &                                 & VGG-S & VGG-Variate \\ \midrule
\multirow{2}{*}{CIFAR-100}     & SCA (Ours)             & 72.2  & 70.3       \\
                               & Full-precision network & 72.2  & 70.4        \\ \midrule
\multirow{2}{*}{Tiny ImageNet} & SCA (Ours)           & 55.2  & 52.1       \\
                               & Full-precision network & 55.6  & 52.4        \\ \bottomrule
\end{tabular}
}
  \end{minipage}
  \hspace{0.1cm}
  \begin{minipage}[t]{0.43\textwidth}
  \centering
        \makeatletter\def\@captype{table}\makeatother\caption{Comparison between SCA and SCA without parameterization on CIFAR-10}
        \label{m4}
        \resizebox{0.99\textwidth}{!}{%
\begin{tabular}{lccc}
\toprule
          & VGG-S &ResNet-20 &VGG-Variate  \\ \midrule
ES$\rm A_{w}$        &  10.00  & 10.00 &  10.00           \\
SCA &\textbf{93.41}     &\textbf{91.28}          &\textbf{92.75}    \\ \bottomrule
\end{tabular}
}
\end{minipage}
\end{table*}

\subsection{Effectiveness of Parameterization with $\tanh()$}
\label{wdrexp}
Parameterization $\tanh(\Theta)$ in SCA is used to compress the full-precision weight space so that we can search for a ternary weight solution more easily.
In this part, we explore the effectiveness of the parameterization.
We compare the performance of SCA with that of SCA without using parameterization $\tanh(\Theta)$ (i.e., the objective function is reduced to $\mathcal{L}(f(X, W), Y)+\lambda*R(W)$).
We denote the latter by ES$\rm A_{w}$.
We use grid search to tune the hyperparameters in ES$\rm A_{w}$.
However, we find that the accuracy of ES$\rm A_{w}$ on CIFAR-10 is always 10\% as there are always some weights that are far from -1, 0, or +1 without the restriction of $\tanh()$.
As shown in Table \ref{m4}, the accuracies drop substantially on CIFAR-10 for all the three networks without parameterization $\tanh()$, which indicates the importance and the effectiveness of the parameterization $\tanh()$ for training ternary weight networks.

\begin{table*}[t]
\centering
\caption{Sparsity (\%) of ternary weights and accuracies  (\%) on MNIST with different values of controller $\alpha$ for given $\lambda$s}
\label{m5}
\begin{tabular}{cccccccccc}
  \toprule
                             &  &  $\alpha$ = 0     & $\alpha$ = 1e-4  &$\alpha$ = 1e-2  &$\alpha$ = 0.1   & $\alpha$ = 0.2   & $\alpha$ = 0.5   &$\alpha$ = 1  &$\alpha$ = 2     \\ \midrule
\multirow{2}{*}{$\lambda$ = 1e-5} & Sparsity & 0.008 & 0.096 & 2.89  & 29.69 & 61.31 & 99.63 & 99.89 & 99.97 \\
                             & Accuracy  & 99.47 & 99.48 & 99.45 & 99.56 & 99.47 & 99.49 & 99.36 & 60.9  \\ \midrule
\multirow{2}{*}{$\lambda$ = 1e-7} & Sparsity & 0.21  & 0.48  & 3.25  & 11.00 & 15.15 & 43.39 & 93.94 & 97.44 \\
                             & Accuracy & 99.53 & 99.56 & 99.40 & 99.48 & 99.50 & 99.50 & 99.40 & 99.42 \\ \midrule
\multirow{2}{*}{$\lambda$ = 1e-9} & Sparsity & 11.48 & 12.65 & 13.41 & 18.30 & 25.86 & 33.46 & 51.99 & 67.21 \\
                             & Accuracy & 99.39 & 99.53 & 99.50 & 99.47 & 99.55 & 99.56 & 99.44 & 99.41 \\ 

    \bottomrule
\end{tabular}
\end{table*}

\begin{table*}[t]
\centering
\setlength{\abovecaptionskip}{0.1cm}
\setlength{\belowcaptionskip}{0.1cm}
\caption{Sparsity (\%) of ternary weights and accuracies (\%) of VGG-S on CIFAR-10 with different values of $\alpha$ for given $\lambda$s}
\label{m6}
\resizebox{1\columnwidth}{!}{%
\begin{tabular}{ccccccccccc}
 \toprule
                           
                             &   &$\alpha$ = 0      &$\alpha$ = 1e-4& $\alpha$ = 1e-2  & $\alpha$ = 0.1  &$\alpha$ = 0.2 &$\alpha$ = 0.5   &$\alpha$ = 1     &$\alpha$ = 1.5   &$\alpha$ = 2     \\ \midrule
\multirow{2}{*}{$\lambda$ = 1e-7} & Sparsity & 0.0143 & 0.099  & 1.001   &2.411       &6.278        & 29.17 & 79.65 & 93.31 & 97.21 \\
                             & Accuracy & 92.77  & 92.66 &  92.42   &92.83     & 92.68      &92.61 & 93.23 & 92.36 & 91.11 \\ \midrule
\multirow{2}{*}{$\lambda$ = 1e-8} & Sparsity & 0.29   & 0.32 & 1.58 & 3.91 &5.38 & 8.56  & 33.92 & 70.28 & 85.77 \\
                             & Accuracy & 93.38  & 93.25 & 93.14 &93.41 &93.00 & 93.29 & 93.15 & 93.36 & 92.75 \\ \midrule
\multirow{2}{*}{$\lambda$ = 1e-9} & Sparsity & 3.02   & 3.05& 4.34  & 7.10 &11.60  & 12.89 & 22.62 & 40.04 & 59.35 \\
                             & Accuracy & 93.05  & 92.90 & 92.97 & 92.96& 92.94 & 93.05 & 92.63 & 93.38 & 92.67 \\ \bottomrule
\end{tabular}
}
\vskip -0.1in
\end{table*} 

\subsection{Sparsity Control}
\label{wdrexp1}
In this part, we investigate how the sparsity (i.e., the percentage of 0s) of the trained ternary weights varies with the values of controller $\alpha$ for given coefficients $\lambda$.
Table \ref{m5} and Table \ref{m6} report the results on MNIST and CIFAR-10, respectively.
It is observed that the ternary weights are becoming increasingly sparser as $\alpha$ increases for given $\lambda$s on both datasets, which verifies that the sparsity of the trained ternary weights is positively related to $\alpha$.
Essentially, when the ternary weights become sparser, the accuracy only changes a little, which indicates that SCA is able to control the sparsity of the trained ternary weights through $\alpha$ with little or no accuracy changes.
For example, on MNIST (Table \ref{m5}), for given $\lambda$ value 1e-5, when $\alpha$ is increased from 0 to 0.5, the sparsity is increased from 0.008\% (which means that there is almost no 0 in the ternary weights) to 99.63\% (which means that there are almost all 0s in the ternary weights), but the accuracy is almost not changed (i.e., from 99.47\% to 99.49\%).
Similarly, on CIFAR-10 (Table \ref{m6}), for given $\lambda$ value 1e-7, when $\alpha$ is increased from 0 to 1, the sparsity is increased from 0.0143\% (which means that there is almost no 0 in the ternary weights) to 79.65\% (which means that most of the values in the ternary weights are 0s), but the accuracy is only changed a little (i.e., from 92.77\% to 93.23\%).
Therefore, SCA can control the sparsity of the trained ternary weights through $\alpha$ with little or no accuracy changes.
We also notice an exception on MNIST that when $\lambda$=1e-5 and $\alpha$=2, the accuracy drops substantially due to over-sparsity with sparsity 99.97\%.
The reason is that 99.97\% 0s make the ternary weight network too sparse to fit the data.

\par

\begin{figure*}[!t]
\centering
     \includegraphics[width=0.98\textwidth]{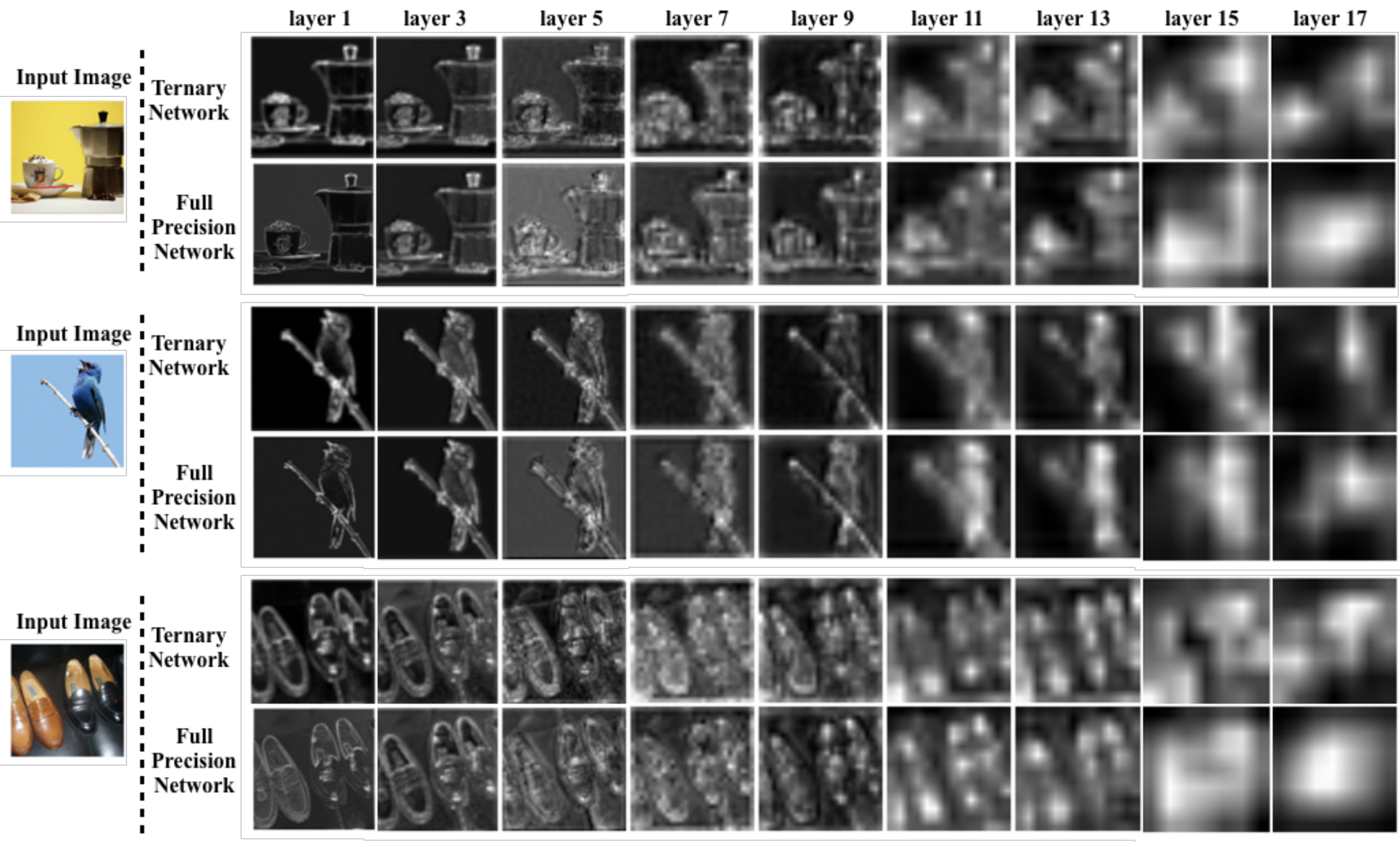}
     \caption{Feature visualization of odd layers of ResNet-18 on ImageNet images}
     \label{f5}
\end{figure*} 

\subsection{Feature Visualization}

The experimental results on several benchmarks datasets have shown that the ternary weight networks trained by SCA perform on a par with the full-precision counterparts.
The natural question that arises is whether their feature maps are similar.
We visualize the average feature map in each odd layer of ternary weight ResNet-18 and full-precision weight ResNet-18.
\par

Figure \ref{f5} shows the feature comparison between each of the odd layers of ternary ResNet-18 and the full-precision ResNet-18 on ImageNet images.
It is observed that their features in the shallow layers (e.g., layer 1-13) are extremely similar and those in the top layers differ a little, which indicates that the ternary weight networks trained by SCA are able to learn effective features similar to those of the full-precision counterparts and thus demonstrates the effectiveness of SCA for feature extraction.

\subsection{Discussion}
We have shown that SCA is able to convert a full-precision network to the ternary version with no or little accuracy decrease.
Besides the performance improvements, the significance of SCA over the existing approaches is more at its simplicity, effectiveness, and sparsity-control ability.
Using a very simple method to achieve even surpass the performances of the state-of-the-art approaches and to control the sparsity of the ternary weights is the main contributions of this work.
From practical perspective, current state-of-the-art models in various applications are built on full-precision weight networks that require large amounts of computation and memory, which highly limits their implementation on resource-limited devices.
SCA paves the way for converting full-precision networks to the ternary versions simply and effectively with a very high code reuse rate, which makes the deployment on resource-limited devices suitable.
Second, SCA is able to control the sparsity of the ternary weights to further accelerate the inference according to the demands on portable devices.

\section{Conclusion}
It is well-known that full-precision DNNs have high-demanding memory and computation requirements. 
In this work, we have proposed a simple yet effective approach SCA to training ternary weight \{-1, 0, +1\} networks that are dramatically efficient at inference and require much less memories.
SCA is able to control the sparsity of the ternary weights, which is to our best knowledge the first work alone this line.
We have theoretically and empirically shown that the sparsity of the ternary weights is positively related to controller $\alpha$.
Extensive experiments on five benchmark datasets have demonstrated the SCA outperforms the existing approaches significantly and even matches the performances of the full-precision counterparts.

\bibliographystyle{elsarticle-num-names}
\bibliography{SCA}

\end{document}